\documentclass{article}

\PassOptionsToPackage{numbers}{natbib}
\usepackage[final]{nips_2017}

\usepackage[utf8]{inputenc} 
\usepackage[T1]{fontenc}    
\usepackage{hyperref}       
\usepackage{url}            
\usepackage{booktabs}       
\usepackage{amsfonts}       
\usepackage{nicefrac}       
\usepackage{microtype}      
\usepackage{amssymb,amsmath}
\usepackage{mathtools}
\mathtoolsset{showonlyrefs}
\usepackage{color}

\usepackage{subfigure} 
\usepackage{algorithm}
\usepackage{algorithmic}
\usepackage{enumitem}
\usepackage{setspace}
\usepackage{multirow}
\usepackage{booktabs}
\usepackage{array}
\usepackage{bm}
\usepackage{wrapfig}

\DeclareMathOperator*{\argmin}{argmin}
\DeclareMathOperator*{\argmax}{argmax}

\DeclareMathOperator*{\sign}{sign}
\DeclareMathOperator*{\diag}{diag}

\DeclareMathOperator*{\vect}{vec}

\DeclareMathOperator*{\conv}{conv}

\newtheorem{proposition}{Proposition} 
 
\newtheorem{theorem}{Theorem} 

\newcommand\cf{cf.\ }
\newcommand\bs[1]{\bm{#1}}

\newcommand\tr{\mathrm{T}}
\newcommand\RHm{\mathbb{R}^{|\mathcal{H}| \times m}}
\newcommand\CfD{C_{f, \mathcal{D}}}
\newcommand\iter[2]{\boldsymbol{#1}^{(#2)}}

\definecolor{supgray}{gray}{0.5}

\newcolumntype{C}[1]{>{\centering\let\newline\\\arraybackslash\hspace{0pt}}m{#1}}
\newcolumntype{A}{ >{$} r <{$} @{} >{${}} l <{$} }

\title{Multi-output Polynomial Networks\\and Factorization Machines}

\author{
  Mathieu Blondel\\
  NTT Communication Science Laboratories \\
  Kyoto, Japan\\
  \texttt{mathieu@mblondel.org} \\
  \And
  Vlad Niculae\thanks{Work performed during an internship at NTT Commmunication
  Science Laboratories, Kyoto.} \\
  Cornell University \\
  Ithaca, NY \\
  \texttt{vlad@cs.cornell.edu} \\
  \And
  Takuma Otsuka \\
  NTT Communication Science Laboratories \\
  Kyoto, Japan\\
  \texttt{otsuka.takuma@lab.ntt.co.jp} \\
  \And
  Naonori Ueda \\
  NTT Communication Science Laboratories \\
  RIKEN \\
  Kyoto, Japan\\
  \texttt{ueda.naonori@lab.ntt.co.jp} \\
}

\begin{document}

\maketitle

\begin{abstract}
Factorization machines and polynomial networks are supervised polynomial models
based on an efficient low-rank decomposition. We extend these models to the
multi-output setting, i.e., for learning vector-valued functions, with
application to multi-class or multi-task problems. We cast this as the problem
of learning a 3-way tensor whose slices share a common basis and propose
a convex formulation of that problem. We then develop an efficient conditional
gradient algorithm and prove its global convergence, despite the fact that it
involves a non-convex basis selection step. On classification tasks, we
show that our algorithm achieves excellent accuracy with much sparser models
than existing methods. On recommendation system tasks, we show how to combine
our algorithm with a reduction from ordinal regression to multi-output
classification and show that the resulting algorithm outperforms simple
baselines in terms of ranking accuracy.
\end{abstract}

\section{Introduction}

Interactions between features play an important role in many classification and
regression tasks. Classically, such interactions have been leveraged 
either explicitly, by mapping features to their products (as in polynomial
regression), or implicitly, through the use of the kernel trick. While fast
linear model solvers have been engineered for the explicit approach
\cite{low_poly,coffin}, they are typically limited to small numbers of features
or low-order feature interactions, due to the fact that the number of parameters
that they need to learn scales as $O(d^t)$, where $d$ is the number of features
and $t$ is the order of interactions considered. Models kernelized with the
polynomial kernel do not suffer from this problem; however, the cost of storing
and evaluating these models grows linearly with the number of training
instances, a problem sometimes referred to as the curse of kernelization
\cite{budgetpegasos}. 

Factorization machines (FMs) \cite{fm} are a more recent approach that can use
pairwise feature interactions efficiently even in very high-dimensional data.
The key idea of FMs is to model the weights of feature interactions using a
\textbf{low-rank} matrix. Not only this idea offers clear benefits in terms of
model compression compared to the aforementioned approaches,
it has also proved instrumental in modeling
interactions between \textbf{categorical} variables, converted to binary
features via a one-hot encoding. Such binary features are usually so sparse that
many interactions are never observed in the training set, preventing classical
approaches from capturing their relative importance. By imposing a low rank on
the feature interaction weight matrix, FMs encourage shared parameters between
interactions, allowing to estimate their weights even if they never occurred in
the training set.  This property has been used in recommender systems to model
interactions between user variables and item variables, and is the basis of
several industrial successes of FMs \cite{fm_param_server,ffm}.

Originally motivated as neural networks with a polynomial activation (instead of
the classical sigmoidal or rectifier activations), polynomial networks (PNs)
\cite{livni} have been shown to be intimately related to FMs and to only subtly
differ in the non-linearity they use \cite{fm_icml}.  PNs 
achieve better performance than rectifier networks on pedestrian
detection \cite{livni} and on dependency parsing \cite{dependency_parsing}, and
outperform kernel approximations such as the Nystr\"{o}m
method \cite{fm_icml}. 
However, existing PN and FM works have been limited
to single-output models, i.e., they are designed to learn scalar-valued
functions, which restricts them to regression or binary classification problems.

\textbf{Our contributions.} In this paper, we generalize FMs and PNs to
multi-output models, i.e., for learning vector-valued functions, with
application to multi-class or multi-task problems.

1) We cast learning multi-output FMs and PNs as learning a 3-way tensor, whose
slices \textbf{share a common basis} (each slice corresponds to one output). To
obtain a \textbf{convex formulation} of that problem, we propose to cast it as
learning an infinite-dimensional but \textbf{row-wise sparse} matrix. 
This can be achieved by using group-sparsity
inducing penalties.  (\S\ref{sec:convex_formulation})

2) To solve the obtained optimization problem, we develop a variant of the
\textbf{conditional gradient} (a.k.a.  Frank-Wolfe) algorithm
\cite{dunn,revisiting_fw}, which repeats the following two steps: i) select a
new basis vector to add to the model and ii) refit the model over the current
basis vectors. (\S\ref{sec:conditional_gradient}) We prove the 
\textbf{global convergence} of this algorithm (Theorem
\ref{theorem:convergence}), despite the fact that the basis selection step is
non-convex and more challenging in the shared basis setting.
(\S\ref{sec:analysis}) 

3) On multi-class classification tasks, we show that our algorithm achieves
comparable accuracy to kernel SVMs but with much more compressed models than the
Nystr\"{o}m method. On recommender system tasks, where kernelized models cannot
be used (since they do not generalize to unseen user-item pairs), we demonstrate
how our algorithm can be combined with a reduction from ordinal regression to
multi-output classification and show that the resulting algorithm
\textbf{outperforms single-output PNs and FMs} both in terms of root
mean squared error (RMSE) and \textbf{ranking
accuracy}, as measured by nDCG (normalized discounted cumulative gain) scores.
(\S\ref{sec:exp})

\vspace{-0.2cm}
\section{Background and related work}

\textbf{Notation.} We denote the set $\{1,\dots,m\}$ by $[m]$.  Given a vector
$\bs{v} \in \mathbb{R}^k$, we denote its elements by $v_r \in \mathbb{R}
~\forall r \in [k]$. Given a matrix $\bs{V} \in \mathbb{R}^{k \times m}$, we
denote its rows by $\bs{v}_r \in \mathbb{R}^m ~\forall r \in [k]$ and its
columns by $\bs{v}_{:,c} ~\forall c \in [m]$. We denote the $l_p$ norm of
$\bs{V}$ by $\|\bs{V}\|_p
\coloneqq \|\vect(\bs{V})\|_p$ and its $l_p/l_q$ norm by $\|\bs{V}\|_{p,q}
\coloneqq \left(\sum_{r=1}^k \|\bs{v}_r\|_q^p\right)^{\frac{1}{p}}$. 
The number of non-zero rows of $\bs{V}$ is denoted by $\|\bs{V}\|_{0,\infty}$.

{\bf Factorization machines (FMs).} Given an input vector 
$\bs{x} \in \mathbb{R}^d$, FMs predict a scalar output by
\begin{equation}
\hat{y}_{\text{FM}} \coloneqq \bs{w}^\tr \bs{x} + \sum_{i < j} w_{i,j} x_i x_j,
\end{equation}
where $\bs{w} \in \mathbb{R}^d$ contains feature weights and $\bs{W} \in
\mathbb{R}^{d \times d}$ is a low-rank matrix that contains pairwise feature
interaction weights. To obtain a low-rank $\bs{W}$, 
\cite{fm} originally proposed to use
a change of variable $\bs{W} = \bs{H}^\tr \bs{H}$, where $\bs{H} \in
\mathbb{R}^{k \times d}$ (with $k \in \mathbb{N}_+$ a rank parameter) and to
learn $\bs{H}$ instead. Noting that this quadratic model results in a non-convex
problem in $\bs{H}$, \cite{convex_fm,cfm_yamada} proposed to convexify the
problem by learning $\bs{W}$ directly but to encourage low rank using a nuclear
norm on $\bs{W}$.  For learning, \cite{convex_fm} proposed a conditional gradient like
approach with global convergence guarantees.

{\bf Polynomial networks (PNs).} PNs are a recently-proposed form of neural
network where the usual activation function is replaced with a squared
activation.  Formally, PNs predict a scalar output by
\begin{equation}
\hat{y}_{\text{PN}} \coloneqq
\bs{w}^\tr \bs{x} +
\bs{v}^\tr \sigma(\bs{H} \bs{x}) 
= \bs{w}^\tr \bs{x} +
\sum_{r=1}^k v_r ~ \sigma(\bs{h}_r^\tr \bs{x}),
\label{eq:polynet}
\end{equation}
where $\sigma(a)=a^2$ (evaluated element-wise) is the squared activation,
$\bs{v} \in \mathbb{R}^k$ is the output layer vector, $\bs{H} \in \mathbb{R}^{k
\times d}$ is the hidden layer matrix and $k$ is the number of hidden units.
Because the r.h.s term can be rewritten as $\bs{x}^\tr \bs{W} \bs{x} =
\sum_{i,j=1}^d w_{i,j} x_i x_j$ if we set $\bs{W} = \bs{H}^\tr \diag(\bs{v})
\bs{H}$, we see that PNs are clearly
a slight variation of FMs and that learning $(\bs{v},\bs{H})$ can be recast as
learning a low-rank matrix $\bs{W}$.  Based on this observation, \cite{livni}
proposed to use GECO \cite{geco}, a greedy algorithm for convex optimization
with a low-rank constraint, similar to the conditional gradient algorithm.
\cite{pn_global} proposed a learning algorithm for PNs with global optimality
guarantees but their theory imposes non-negativity on the network parameters and
they need one distinct hyper-parameter per hidden unit to avoid trivial models.
Other low-rank polynomial models were recently introduced in
\cite{sl_tensor_net,exp_machines} but using a tensor network (a.k.a. tensor
train) instead of the canonical polyadic (CP) decomposition.

\section{A convex formulation of multi-output PNs and FMs}
\label{sec:convex_formulation}

\begin{figure}[t]
\centering
\includegraphics[scale=0.32]{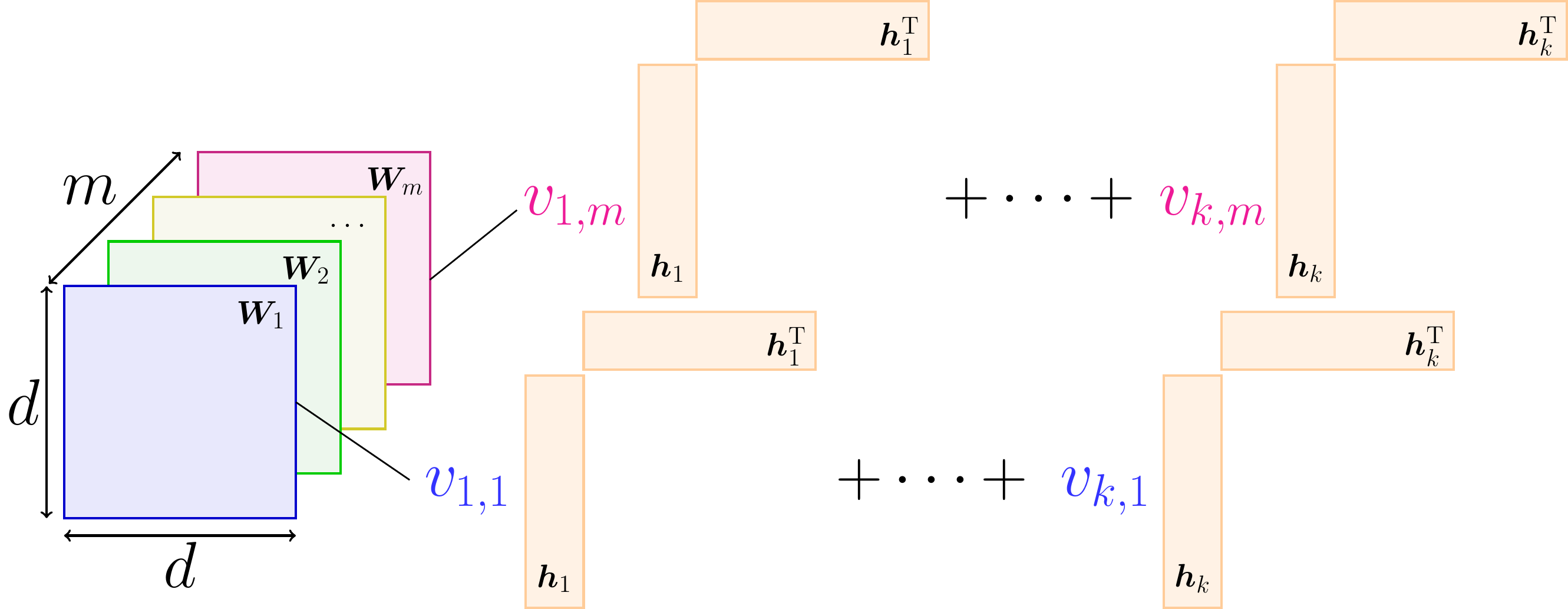}
\caption{Our multi-output PNs / FMs learn a tensor whose slices
    \textbf{share a common basis} $\{\bs{h}_r\}_{r=1}^k$.}
\end{figure}
In this section, we generalize PNs and FMs to multi-output problems. For the
sake of concreteness, we focus on PNs for multi-class classification. The
extension to FMs is straightforward and simply requires to replace
$\sigma(\bs{h}^\tr \bs{x}) = (\bs{h}^\tr \bs{x})^2$ by
$\sigma_\text{ANOVA}(\bs{h},\bs{x}) \coloneqq \sum_{i < j} x_i h_i x_j h_j$,
as noted in \cite{fm_icml}.

The predictions of multi-class PNs can be naturally defined as
$\hat{y}_{\text{MPN}} \coloneqq \argmax_{c \in [m]} \bs{w}_c^\tr \bs{x} +
\bs{x}^\tr \bs{W}_c \bs{x}$, where $m$ is the number of classes, $\bs{w}_c \in
\mathbb{R}^d$ and $\bs{W}_c \in \mathbb{R}^{d \times d}$ is low-rank.
Following \cite{fm_icml}, we can model the linear term directly in the quadratic
term if we augment all data points with an extra feature of value 1, i.e.,
$\bs{x}^\tr \leftarrow [1, \bs{x}^\tr]$. We will therefore simply assume
$\hat{y}_{\text{MPN}} = \argmax_{c \in [m]} \bs{x}^\tr \bs{W}_c
\bs{x}$ henceforth. Our main proposal in this paper is to decompose $\bs{W}_1,
\dots, \bs{W}_m$ using a \textbf{shared} basis:
\begin{equation}
\bs{W}_c = \bs{H}^\tr \diag(\bs{v}_{:,c}) \bs{H} = \textstyle{\sum_{r=1}^k
v_{r,c} \bs{h}_r \bs{h}_r^\tr} \quad \forall c \in [m],
\label{eq:Wc_decomp}
\end{equation}
where, in neural network terminology, $\bs{H} \in \mathbb{R}^{k \times d}$ can
be interpreted as a hidden layer matrix and $\bs{V} \in \mathbb{R}^{k \times m}$
as an output layer matrix.  Compared to the naive approach of decomposing each
$\bs{W}_c$ as $\bs{W}_c = \bs{H}^\tr_c \diag(\bs{v}_{:,c}) \bs{H}_c$, this
reduces the number of parameters from $m(dk+k)$ to $dk+mk$. 

While a nuclear norm could be used to promote a low rank on each $\bs{W}_c$,
similarly as in \cite{convex_fm,cfm_yamada}, this is clearly not sufficient to
impose a shared basis. A naive approach would be to use non-orthogonal
joint diagonalization as a post-processing. However, because this is a
non-convex problem for which no globally convergent algorithm is known
\cite{beyond_cca}, this would result in a loss of accuracy. Our key idea is to
cast the problem of learning a multi-output PN as that of learning an
\textbf{infinite but row-wise sparse matrix}.  Without loss of generality, we
assume that basis vectors (hidden units) lie in the unit ball. We therefore
denote the set of basis vectors by
$
\mathcal{H} \coloneqq \{\bs{h} \in \mathbb{R}^d \colon \|\bs{h}\|_2 \le 1\}.
$
Let us denote this infinite matrix by $\bs{U} \in \RHm$ (we use a discrete
notation for simplicity).  We can then write 
\begin{equation}
\hat{y}_{\text{MPN}} = \argmax_{c \in [m]} ~
\bs{o}(\bs{x}; \bs{U})_c
\quad \text{where} \quad
\bs{o}(\bs{x}; \bs{U}) \coloneqq
\sum_{\bs{h} \in \mathcal{H}} ~ \sigma(\bs{h}^\tr \bs{x})
\bs{u}_{\bs{h}} \in \mathbb{R}^m
\quad \text{and}
\end{equation}
$\bs{u}_{\bs{h}} \in \mathbb{R}^m$ denotes the
weights of basis $\bs{h}$ across all classes (outputs). In this
formulation, we have $\bs{W}_c = \sum_{\bs{h} \in \mathcal{H}} u_{\bs{h},c}
\bs{h} \bs{h}^\tr$ and sharing a common basis (hidden units) amounts to
encouraging the rows of $\bs{U}$, $\bs{u}_{\bs{h}}$, to be either dense or entirely sparse.  
This can be naturally achieved using group-sparsity inducing penalties.
Intuitively, $\bs{V}$ in \eqref{eq:Wc_decomp} can be thought as $\bs{U}$
restricted to its row support.  
Define the training set by $\bs{X} \in
\mathbb{R}^{n \times d}$  and $\bs{y} \in [m]^n$.  We
then propose to solve the convex problem
\begin{equation}
\min_{\Omega(\bs{U}) \le \tau} F(\bs{U})
\coloneqq \sum_{i=1}^n \ell\left(y_i, 
\bs{o}(\bs{x}_i; \bs{U})
\right),
\label{eq:convex_obj}
\end{equation}
where $\ell$ is a smooth and convex multi-class loss function (\cf Appendix
\ref{appendix:loss_func} for three common examples), $\Omega$ is a
sparsity-inducing penalty and $\tau > 0$ is a hyper-parameter. In this paper, we
focus on the $l_1$ (lasso), $l_1/l_2$ (group lasso) and $l_1/l_\infty$ penalties
for $\Omega$, \cf Table \ref{table:norms}.  However, as we shall see, solving
\eqref{eq:convex_obj} is more challenging with the $l_1/l_2$
and $l_1/l_\infty$ penalties than with the $l_1$ penalty.
Although our formulation is based on an infinite view, we next show that
$\bs{U}^\star$ has finite row support.  
\begin{proposition}{Finite row support of $\bs{U}^\star$ for multi-output PNs
    and FMs}

Let $\bs{U}^\star$ be an optimal solution of \eqref{eq:convex_obj}, where
$\Omega$ is one of the penalties in Table \ref{table:norms}. Then,\\
$\|\bs{U}^\star\|_{0,\infty} \le nm+1$.  If $\Omega(\cdot)=\|\cdot\|_1$, we can
tighten this bound to $\|\bs{U}^\star\|_{0,\infty} \le \min(nm+1,dm)$.
\label{proposition:support}
\end{proposition}
Proof is in Appendix \ref{appendix:support}. It is open
whether we can tighten this result when $\Omega=\|\cdot\|_{1,2}$ or
$\|\cdot\|_{1,\infty}$.
\begin{table*}[t]
\caption{Sparsity-inducing penalties considered in this paper.  With some abuse of
    notation, we denote by $\bs{e}_{\bs{h}}$ and $\bs{e}_c$ standard basis
    vectors of dimension $|\mathcal{H}|$ and $m$, respectively.  Selecting an
    optimal basis vector $\bs{h}^\star$ to add is a non-convex optimization
    problem.  The constant $\epsilon \in (0,1)$ is the tolerance parameter used
    for the power method and $\nu$ is the multiplicative approximation we
    guarantee. 
}
\label{table:norms}
\begin{center}
\begin{scriptsize}
\begin{tabular}{l l l c A c}
\toprule
& $\Omega(\bs{U})$ & $\Omega^*(\bs{G})$ & \
$\bs{\Delta}^\star \in \tau \cdot \partial \Omega^*(\bs{G})$ &
\multicolumn{2}{c}{Subproblem} & $\nu$ \\
\midrule
\addlinespace[0.5em]
$l_1$ (lasso) & $\|\bs{U}\|_1$ & $\|\bs{G}\|_\infty$ &
$\tau \sign(g_{\bs{h}^\star,c^\star}) \bs{e}_{\bs{h}^\star} \bs{e}_{c^\star}^\tr$ &
\bs{h}^\star, c^\star &\in \displaystyle{\argmax_{\bs{h} \in \mathcal{H}, c \in [m]}}
|g_{\bs{h},c}| & $1-\epsilon$ \\
\addlinespace[0.7em]

$l_1/l_2$ (group lasso) & $\|\bs{U}\|_{1,2}$ & $\|\bs{G}\|_{\infty,2}$ &
$\tau \bs{e}_{\bs{h}^\star} \bs{g}_{\bs{h}^\star}^\tr / \|\bs{g}_{\bs{h}^\star}\|_2$ &
\bs{h}^\star &\in \displaystyle{\argmax_{\bs{h} \in \mathcal{H}}}
\|\bs{g}_{\bs{h}}\|_2 & $\frac{1-\epsilon}{\sqrt{m}}$ \\
\addlinespace[0.7em]

$l_1/l_\infty$ & $\|\bs{U}\|_{1,\infty}$ & $\|\bs{G}\|_{\infty,1}$ &
$\tau \bs{e}_{\bs{h}^\star} \sign(\bs{g}_{\bs{h}^\star})^\tr$ &
\bs{h}^\star &\in \displaystyle{\argmax_{\bs{h} \in \mathcal{H}}}
\|\bs{g}_{\bs{h}}\|_1  & $\frac{1-\epsilon}{m}$ \\
\bottomrule
\end{tabular}
\end{scriptsize}
\end{center}
\end{table*}

\section{A conditional gradient algorithm with approximate basis vector
selection}
\label{sec:conditional_gradient}

At first glance, learning with an infinite number of basis vectors seems
impossible. In this section, we show how the well-known conditional gradient
algorithm \cite{dunn,revisiting_fw} combined with group-sparsity inducing
penalties naturally leads to a greedy algorithm that \textbf{selects and adds
basis vectors that are useful across all outputs}.  On every iteration, the
conditional gradient algorithm performs updates of the form $\iter{U}{t+1} = (1
- \gamma) \iter{U}{t} + \gamma \bs{\Delta}^\star$, where $\gamma \in [0, 1]$ is
a step size and $\bs{\Delta}^\star$ is obtained by solving a linear
approximation of the objective around the current iterate $\iter{U}{t}$:
\begin{equation}
    \bs{\Delta}^\star \in \argmin_{\Omega(\bs{\Delta}) \le \tau}
    \langle \bs{\Delta}, \nabla F(\iter{U}{t}) \rangle
    = \tau \cdot \argmax_{\Omega(\bs{\Delta}) \le 1}
\langle \bs{\Delta}, -\nabla F(\iter{U}{t}) \rangle.
\label{eq:linearized_pb}
\end{equation}
Let us denote the negative gradient $-\nabla F(\bs{U})$ by
$\bs{G} \in \RHm$ for short. Its elements are defined by
\begin{equation}
g_{\bs{h},c} 
= -\sum_{i=1}^n 
\sigma(\bs{h}^\tr \bs{x}_i)
\nabla \ell\left(y_i, 
\bs{o}(\bs{x}_i; \bs{U})
\right)_c,
\label{eq:Gamma_general_case}
\end{equation}
where $\nabla \ell(y, \bs{o}) \in \mathbb{R}^m$ is the gradient of $\ell$ w.r.t.
$\bs{o}$ (\cf Appendix \ref{appendix:loss_func}).
For ReLu activations, solving \eqref{eq:linearized_pb} is
known to be NP-hard \cite{convex_nn_bach}.  Here, we focus on quadratic
activations, for which we will be able to provide approximation guarantees.
Plugging the expression of $\sigma$, we get
\begin{equation}
g_{\bs{h},c} = -\bs{h}^\tr \bs{\Gamma}_c \bs{h} 
~ \text{where} ~
\bs{\Gamma}_c \coloneqq \bs{X}^\tr \bs{D}_c \bs{X} \text{ (PN) or }
\bs{\Gamma}_c \coloneqq \frac{1}{2} \Big(\bs{X}^\tr \bs{D}_c \bs{X} - 
    \bs{D}_c \sum_{i=1}^n \diag(\bs{x}_i)^2\Big) \text{ (FM)}
\end{equation}
and $\bs{D}_c \in \mathbb{R}^{n \times n}$ is a diagonal matrix such
that $(\bs{D}_c)_{i,i} \coloneqq \nabla \ell(y_i, 
\bs{o}(\bs{x}_i; \bs{U}))_c$.
Let us recall the definition of the dual norm of $\Omega$:
$\Omega^*(\bs{G}) \coloneqq \max_{\Omega(\bs{\Delta}) \le 1}
\langle \bs{\Delta}, \bs{G} \rangle$.
By comparing this equation to \eqref{eq:linearized_pb}, we see that
$\bs{\Delta}^\star$ is the argument that achieves the maximum in the dual norm
$\Omega^*(\bs{G})$, up to a constant factor $\tau$.  It is easy to verify that
any element in the subdifferential of $\Omega^*(\bs{G})$, which we denote by
$\partial \Omega^*(\bs{G}) \subseteq \RHm$, achieves that maximum, i.e.,
$\bs{\Delta}^\star \in \tau \cdot \partial \Omega^*(\bs{G})$.

\textbf{Basis selection.} As shown in Table \ref{table:norms},
elements of $\partial \Omega^*(\bs{G})$ (subgradients) are $|\mathcal{H}| \times
m$ matrices with a \textbf{single non-zero row} indexed by $\bs{h}^\star$, where
$\bs{h}^\star$ is an optimal basis (hidden unit) selected by
\begin{equation}
    \bs{h}^\star \in \argmax_{\bs{h} \in \mathcal{H}} \|\bs{g}_{\bs{h}}\|_p,
\label{eq:unit_selection}
\end{equation}
and where $p=\infty$ when $\Omega=\|\cdot\|_1$, $p=2$ when $\Omega=\|.\|_{1,2}$
and $p=1$ when $\Omega=\|\cdot\|_{1,\infty}$. We call \eqref{eq:unit_selection}
a basis vector selection criterion.  Although this selection criterion was
derived from the linearization of the objective, it is fairly natural: it
chooses the basis vector with largest ``violation'', as measured by
the $l_p$ norm of the negative gradient row $\bs{g}_{\bs{h}}$.

{\bf Multiplicative approximations.}
The key challenge in solving \eqref{eq:linearized_pb} or equivalently
\eqref{eq:unit_selection} arises from the fact that $\bs{G}$ has infinitely many
rows $\bs{g}_{\bs{h}}$. We therefore cast basis vector selection as a continuous
optimization problem w.r.t. $\bs{h}$. Surprisingly, although the entire
objective \eqref{eq:convex_obj} is convex, \eqref{eq:unit_selection} is not. 
Instead of the exact maximum, we will therefore only require to find a
$\bs{\hat{\Delta}} \in \RHm$ that satisfies
\begin{equation}
    \Omega(\bs{\hat{\Delta}}) \le \tau
    \quad \text{and} \quad
    \langle \bs{\hat{\Delta}}, \bs{G} \rangle \ge \nu 
    \langle \bs{\Delta}^\star, \bs{G} \rangle,
\end{equation}
where $\nu \in (0,1]$ is a multiplicative approximation (higher is better).  It
is easy to verify that this is equivalent to replacing the optimal
$\bs{h}^\star$ by an approximate $\bs{\hat{h}} \in \mathcal{H}$ that satisfies
$\|\bs{g}_{\bs{\hat{h}}}\|_p \ge \nu \|\bs{g}_{\bs{h}^\star}\|_p$.

{\bf Sparse case.} When $\Omega(\cdot)=\|\cdot\|_1$, we need to solve
\begin{equation}
\max_{\bs{h} \in \mathcal{H}} \|\bs{g}_{\bs{h}}\|_\infty =
\max_{\bs{h} \in \mathcal{H}} \max_{c \in [m]} |\bs{h}^\tr
\bs{\Gamma}_c \bs{h}| = \max_{c \in [m]} \max_{\bs{h} \in \mathcal{H}}
|\bs{h}^\tr \bs{\Gamma}_c \bs{h}|.
\end{equation}
It is well known that the optimal solution of $\max_{\bs{h} \in \mathcal{H}}
|\bs{h}^\tr \bs{\Gamma}_c \bs{h}|$ is the dominant eigenvector of
$\bs{\Gamma}_c$.  Therefore, we simply need to find the dominant eigenvector
$\bs{h}_c$ of each $\bs{\Gamma}_c$ and select $\bs{\hat{h}}$ as the $\bs{h}_c$
with largest singular value $|\bs{h}_c^\tr \bs{\Gamma}_c \bs{h}_c|$.  Using the
power method, we can find an $\bs{h}_c$ that satisfies
\begin{equation}
|\bs{h}_c^\tr \bs{\Gamma}_c \bs{h}_c| \ge (1 - \epsilon)
\max_{\bs{h} \in \mathcal{H}} |\bs{h}^\tr \bs{\Gamma}_c \bs{h}|,
\label{eq:power_method}
\end{equation}
for some tolerance parameter $\epsilon \in (0, 1)$.  The procedure takes
$\mathcal{O}(N_c \log(d)/\epsilon)$ time, where $N_c$ is the number of non-zero
elements in $\bs{\Gamma}_c$ \cite{geco}. Taking the maximum w.r.t. $c \in [m]$
on both sides of \eqref{eq:power_method} leads to
$\|\bs{g}_{\bs{\hat{h}}}\|_\infty \ge \nu \|\bs{g}_{\bs{h}^\star}\|_\infty$,
where $\nu = 1-\epsilon$. However, using $\Omega=\|\cdot\|_1$ does not encourage
selecting an $\bs{\hat{h}}$ that is useful for all outputs. In fact, when
$\Omega=\|\cdot\|_1$, our approach is equivalent to imposing
independent nuclear norms on $\bs{W}_1,\dots,\bs{W}_m$.

{\bf Group-sparse cases.} When
$\Omega(\cdot)=\|.\|_{1,2}$ or $\Omega(\cdot)=\|.\|_{1,\infty}$,
we need to solve
\begin{equation}
\max_{\bs{h} \in \mathcal{H}} \|\bs{g}_h\|_2^2 
= \max_{\bs{h} \in \mathcal{H}} f_2(\bs{h}) \coloneqq \sum_{c=1}^m
(\bs{h}^\tr \bs{\Gamma}_c \bs{h})^2 \quad \text{or} \quad
\max_{\bs{h} \in \mathcal{H}} \|\bs{g}_h\|_1 
= \max_{\bs{h} \in \mathcal{H}} f_1(\bs{h}) \coloneqq \sum_{c=1}^m
|\bs{h}^\tr \bs{\Gamma}_c \bs{h}|,
\end{equation}
respectively. Unlike the $l_1$-constrained case, we are clearly selecting a
basis vector with largest violation \textbf{across all outputs}. However, we are
now faced with a more difficult non-convex optimization problem. Our strategy is
to first choose an initialization $\iter{h}{0}$ which guarantees a certain
multiplicative approximation $\nu$, then refine the solution using a
monotonically non-increasing iterative procedure.

\emph{Initialization.} We simply choose $\iter{h}{0}$ as the approximate
solution of the $\Omega=\|\cdot\|_1$ case, i.e., we have
\begin{equation}
\|\bs{g}_{\iter{h}{0}}\|_\infty \ge (1-\epsilon)
\max_{\bs{h} \in \mathcal{H}} \|\bs{g}_{\bs{h}}\|_\infty.
\end{equation}
Now, using 
$\sqrt{m} \|\bs{x}\|_\infty \ge \|\bs{x}\|_2 \ge \|\bs{x}\|_\infty$ and $m
\|\bs{x}\|_\infty \ge \|\bs{x}\|_1 \ge \|\bs{x}\|_\infty$, this immediately
implies
\begin{equation}
\|\bs{g}_{\iter{h}{0}}\|_p \ge \nu \max_{\bs{h} \in \mathcal{H}}
\|\bs{g}_{\bs{h}}\|_p,
\end{equation}
with 
$\nu=\frac{1-\epsilon}{\sqrt{m}}$ if $p=2$ and $\nu=\frac{1-\epsilon}{m}$ if
$p=1$.

\emph{Refining the solution.} We now apply another instance of the conditional
gradient algorithm to solve the subproblem $\max_{\|\bs{h}\|_2 \le 1}
f_p(\bs{h})$ itself, leading to the following iterates:
\begin{equation}
\iter{h}{t+1} = (1-\eta_t) \iter{h}{t} + \eta_t \frac{\nabla
f_p(\iter{h}{t})}{\|\nabla f_p(\iter{h}{t})\|_2},
\label{eq:refine_recursion}
\end{equation}
where $\eta_t \in [0,1]$. Following \cite[Section 2.2.2]{bertsekas}, if we use
the Armijo rule to select $\eta_t$, every limit point of the sequence
$\{\iter{h}{t}\}$ is a stationary point of $f_p$. In practice, we observe that
$\eta_t=1$ is almost always selected. Note that when $\eta_t=1$ and $m=1$ (i.e.,
single-output case), our refining algorithm \textbf{recovers the power method}.
Generalized power methods were also studied for structured matrix
factorization \cite{journee_generalized,luss_generalized}, but with
different objectives and constraints.
Since the conditional gradient algorithm assumes a differentiable function, in
the case $p=1$, we replace the absolute function with the Huber function
$|x|\approx \frac{1}{2} x^2$ if $|x|\le 1$, $|x|-\frac{1}{2}$ otherwise. 

\begin{figure*}
\centering
\includegraphics[scale=1.0]{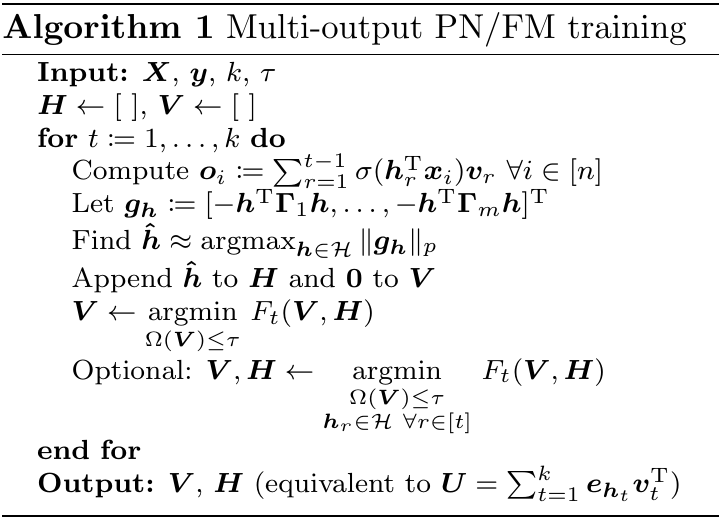}
\end{figure*}

\textbf{Corrective refitting step.} 
After $t$ iterations, $\iter{U}{t}$ contains at most $t$ non-zero rows.
We can therefore always store $\iter{U}{t}$ as $\iter{V}{t} \in
\mathbb{R}^{t \times m}$ (the output layer matrix) and $\iter{H}{t} \in
\mathbb{R}^{t \times d}$ (the basis vectors / hidden units added so far).
In order to improve accuracy, on iteration $t$, we can then
refit the objective $F_t(\bs{V}, \bs{H}) \coloneqq \sum_{i=1}^n \ell\left(y_i,
\sum_{r=1}^t \sigma(\bs{h}_r^\tr \bs{x}_i) \bs{v}_r \right)$. We consider two
kinds of corrective steps, a convex one that minimizes $F_t(\bs{V},
\iter{H}{t})$ w.r.t. $\bs{V} \in \mathbb{R}^{t \times m}$ and an
optional non-convex one that minimizes $F_t(\bs{V}, \bs{H})$ w.r.t. both
$\bs{V} \in \mathbb{R}^{t \times m}$ and $\bs{H} \in \mathbb{R}^{t \times d}$.
Refitting allows to remove previously-added bad basis vectors, thanks to the use
of sparsity-inducing penalties.  Similar refitting procedures are commonly used
in matching pursuit \citep{omp}.  The entire procedure is summarized in
Algorithm 1 and implementation details are given in Appendix
\ref{appendix:impl_detail}.

\section{Analysis of Algorithm 1}
\label{sec:analysis}

The main difficulty in analyzing the convergence of Algorithm 1 stems from the
fact that we cannot solve the basis vector selection subproblem globally when
$\Omega=\|\cdot\|_{1,2}$ or $\|\cdot\|_{1,\infty}$.  Therefore, we need to
develop an analysis that can cope with the multiplicative approximation $\nu$.
Multiplicative approximations were also considered in \cite{block_fw} but the
condition they require is too stringent (\cf Appendix
\ref{appendix:convergence_analysis} for a detailed discussion).  The next
theorem guarantees the number of iterations needed to output a multi-output
network that achieves as small objective value as an optimal solution of
\eqref{eq:convex_obj}.
\begin{theorem}{Convergence of Algorithm 1}

Assume $F$ is smooth with constant $\beta$.  Let $\iter{U}{t}$ be the output
after $t$ iterations of Algorithm 1 run with
constraint parameter $\frac{\tau}{\nu}$.  Then,
$F(\iter{U}{t}) - \displaystyle{\min_{\Omega(\bs{U}) \le \tau}} F(\bs{U}) \le
\epsilon ~\forall t \ge \frac{8 \tau^2 \beta}{\epsilon \nu^2} - 2$. \\
\label{theorem:convergence}
\end{theorem}
In \cite{livni}, single-output PNs were trained using GECO \cite{geco}, a greedy
algorithm with similar $\mathcal{O}\big(\frac{\tau^2 \beta}{\epsilon
\nu^2}\big)$ guarantees. However, GECO is limited to learning infinite vectors
(not matrices) and it does not constrain its iterates like we do.  Hence GECO
cannot remove bad basis vectors.  The proof of Theorem \ref{theorem:convergence}
and a detailed comparison with GECO are given in Appendix
\ref{appendix:convergence_analysis}. Finally, we note that the infinite
dimensional view is also key to convex neural networks \cite{convex_nn_bengio,
convex_nn_bach}. However, to our knowledge, we are the first to give an explicit
multiplicative approximation guarantee for a non-linear multi-output network.

\vspace{-0.2cm}
\section{Experimental results}
\label{sec:exp}

\subsection{Experimental setup}

{\bf Datasets.} For our multi-class experiments, we use four publicly-available
datasets: segment (7 classes), vowel (11 classes), satimage (6 classes) and
letter (26 classes) \cite{datasets}. Quadratic
models substantially improve over linear models on these datasets. For our
recommendation system experiments, we use the MovieLens 100k and 1M datasets
\cite{movielens}. See Appendix \ref{appendix:datasets} for complete details.

{\bf Model validation.} The greedy nature of Algorithm 1 allows us to easily
interleave training with model validation. Concretely, we use an outer loop
(embarrassingly parallel) for iterating over the range of possible
regularization parameters, and an inner loop (Algorithm 1, sequential) for
increasing the number of basis vectors. Throughout our experiments, we use 50\%
of the data for training, 25\% for validation, and 25\% for evaluation.  Unless
otherwise specified, we use a multi-class logistic loss.

\subsection{Method comparison for the basis vector (hidden unit) selection
subproblem}

\begin{wrapfigure}{r}{6.5cm} 
\vspace{-12pt}
\includegraphics[scale=0.30]{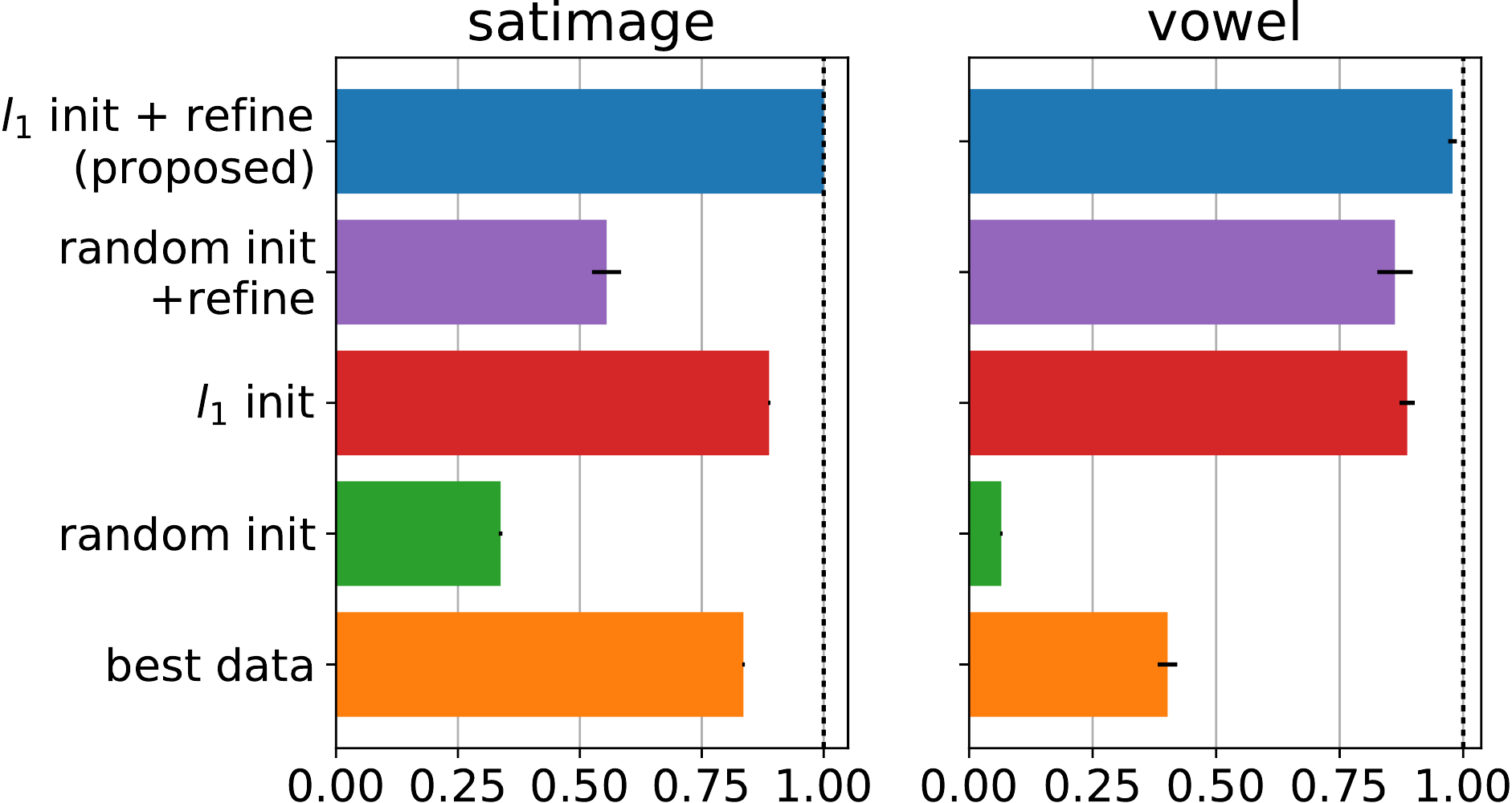}
\caption{ Empirically observed multiplicative approximation factor $\hat{\nu} =
f_1(\bs{\hat{h}}) / f_1(\bs{h}^\star)$.}
\label{fig:subproblem}
\end{wrapfigure}
As we mentioned previously, the linearized subproblem (basis vector selection)
for the $l_1/l_2$ and $l_1/l_\infty$ constrained cases involves a significantly
more challenging non-convex optimization problem. In this section, we compare
different methods for obtaining an approximate solution $\bs{\hat{h}}$ to
\eqref{eq:unit_selection}. We focus
on the $\ell_1/\ell_\infty$ case, since we have a method for computing the true
global solution $\bs{h}^\star$, albeit with exponential complexity in $m$ (\cf
Appendix \ref{appendix:l1linf_subproblem}). This allows us to report the
empirically observed multiplicative approximation factor $\hat{\nu} \coloneqq
f_1(\bs{\hat{h}}) / f_1(\bs{h}^\star)$. 

{\bf Compared methods.} We compare \emph{$l_1$ init + refine} (proposed),
\emph{random init + refine}, \emph{$l_1$ init} (without refine), \emph{random
init} and \emph{best data}: $\bs{\hat{h}} = \bs{x}_{i^\star} /
\|\bs{x}_{i^\star}\|_2$ where $i^\star = \displaystyle{\argmax_{i \in [n]}} ~
f_1(\bs{x}_i / \|\bs{x}_i\|_2)$.

{\bf Results.} We report $\hat{\nu}$ in Figure \ref{fig:subproblem}.
\textit{$l_1$ init + refine} achieves nearly the global maximum on both datasets
and outperforms \textit{random init + refine}, showing the effectiveness of the
proposed initialization and that the iterative update
\eqref{eq:refine_recursion} can get stuck in a bad local minimum if initialized
badly. On the other hand, \textit{$l_1$ init + refine} outperforms \textit{$l_1$
init} alone, showing the importance of iteratively refining the solution.
\textit{Best data}, a heuristic similar to that of approximate kernel SVMs
\cite{lasvm}, is not competitive.

\subsection{Sparsity-inducing penalty comparison}

\begin{wrapfigure}{r}{4.4cm}
\vspace{-15pt}
\subfigure{\includegraphics[width=0.30\textwidth]{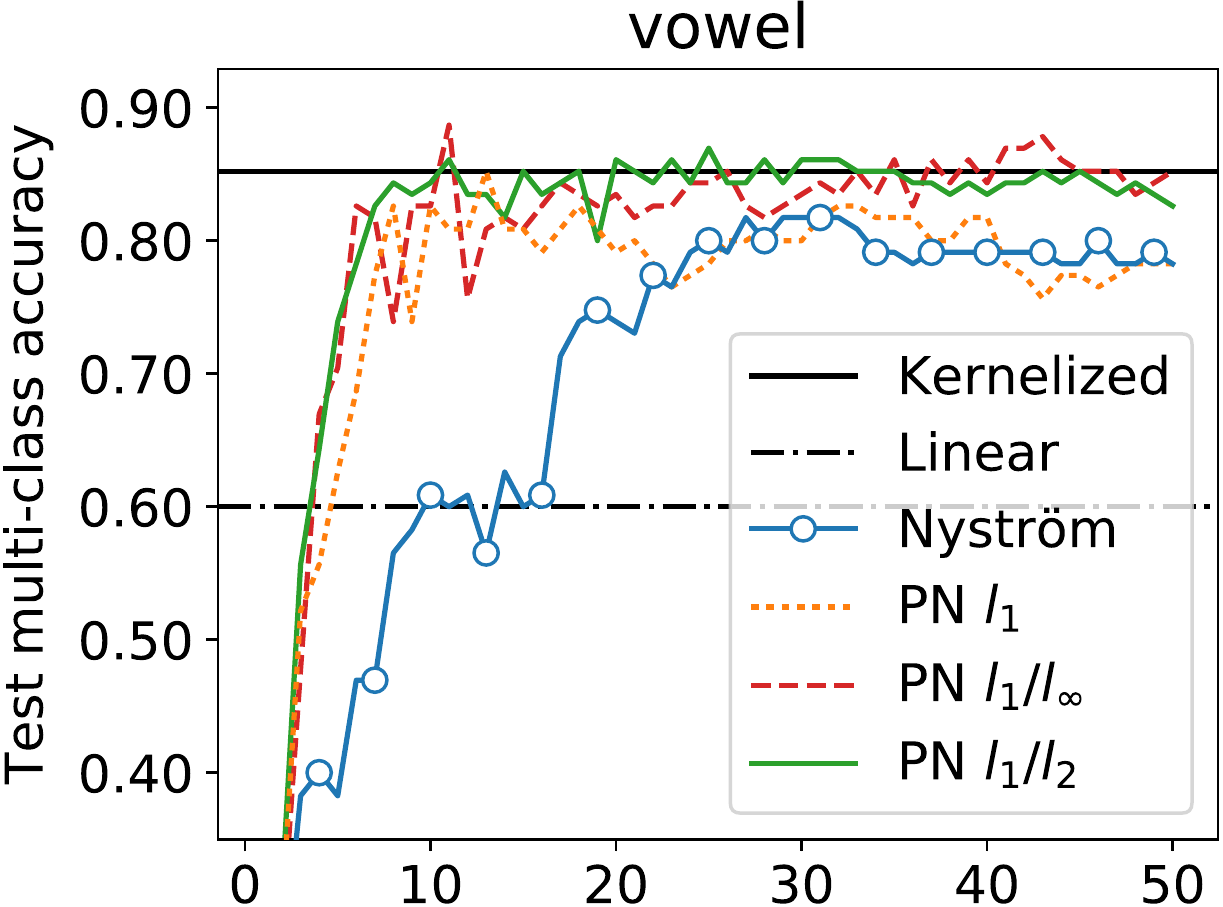}} \\
\subfigure{\includegraphics[width=0.30\textwidth]{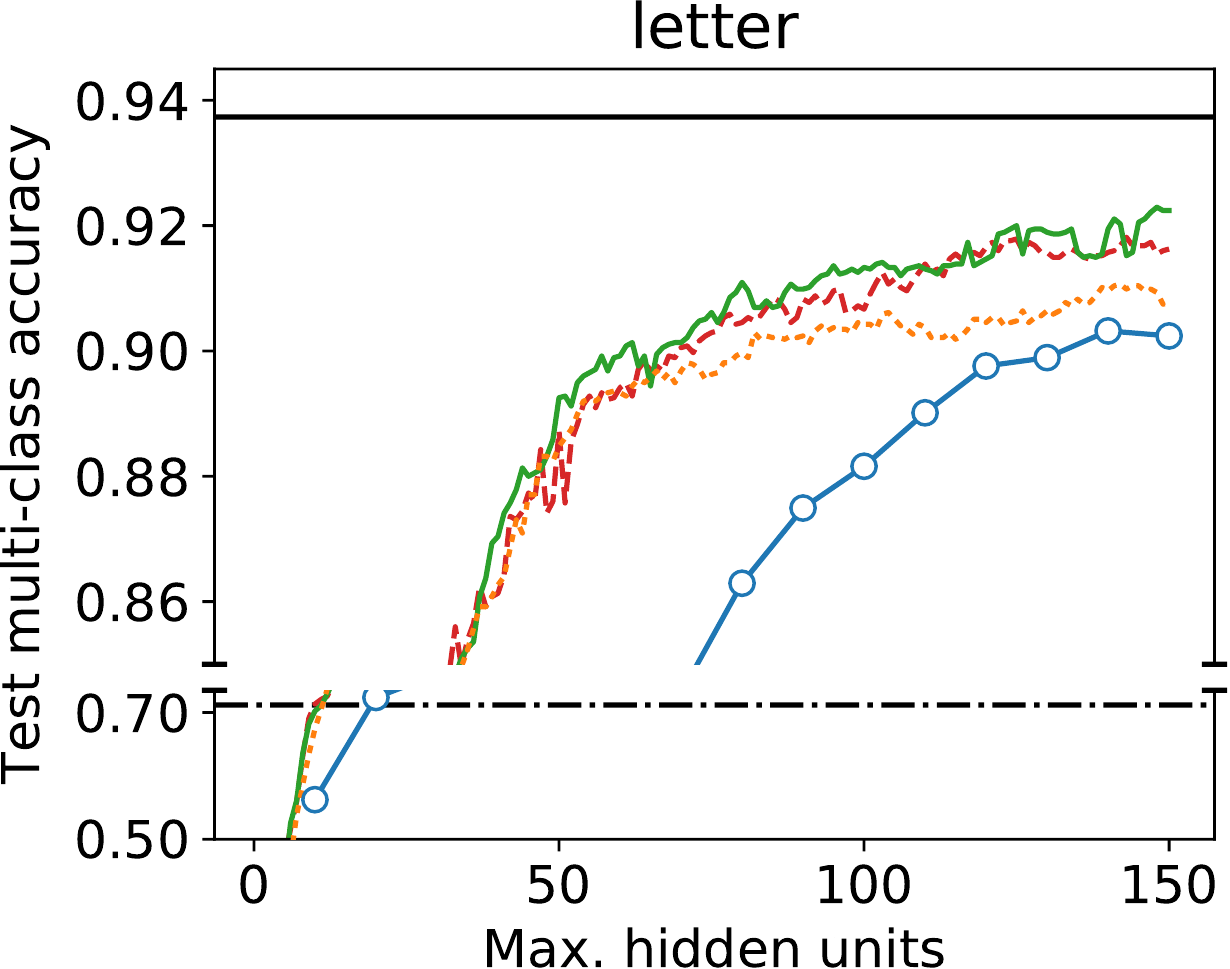}}%
\caption{Penalty comparison.}
\label{fig:penalty}
\end{wrapfigure}
In this section, we compare the $l_1$, $l_1/l_2$ and $l_1/l_\infty$ penalties for
the choice of $\Omega$,  when varying the maximum number of basis vectors
(hidden units). Figure
\ref{fig:penalty} indicates test set accuracy when using output layer refitting.
We also include linear logistic regression, kernel SVMs and the Nystr\"{o}m
method as baselines. For the latter two, we use the quadratic
kernel $(\bs{x}_i^\tr \bs{x}_j + 1)^2$.  Hyper-parameters are chosen so as to
maximize validation set accuracy.\\[0.1cm]
{\bf Results.} On the vowel (11 classes) and letter (26 classes) datasets,
$l_1/l_2$ and $l_1/l_\infty$ penalties outperform $l_1$ norm starting from 20 and 75
hidden units, respectively.  On satimage (6 classes) and segment (7 classes),
we observed that the three penalties are mostly similar (not shown).  We hypothesize
that $l_1/l_2$ and $l_1/l_\infty$ penalties make a bigger difference when the
number of classes is large. Multi-output PNs substantially outperform the
Nystr\"{o}m method with comparable number of basis vectors (hidden units).
Multi-output PNs reach the same test accuracy as kernel SVMs with very few
basis vectors on vowel and satimage but appear to require at least 100 basis
vectors to reach good performance on letter. This is not surprising, since kernel
SVMs require 3,208 support vectors on letter, as indicated in Table
\ref{table:scores} below.

\subsection{Multi-class benchmark comparison}

{\bf Compared methods.} We compare
the proposed conditional gradient algorithm with output layer refitting only and
with both output and hidden layer refitting; projected gradient descent (FISTA)
with random initialization; linear and kernelized models; one-vs-rest PNs (i.e.,
fit one PN per class). We focus on PNs rather than FMs since they are known to
work better on classification tasks \cite{fm_icml}. \\[0.1cm]
\begin{table}[t]
\caption{Muli-class test accuracy and number of basis vectors / support vectors.}
\centering
\includegraphics[scale=1.1]{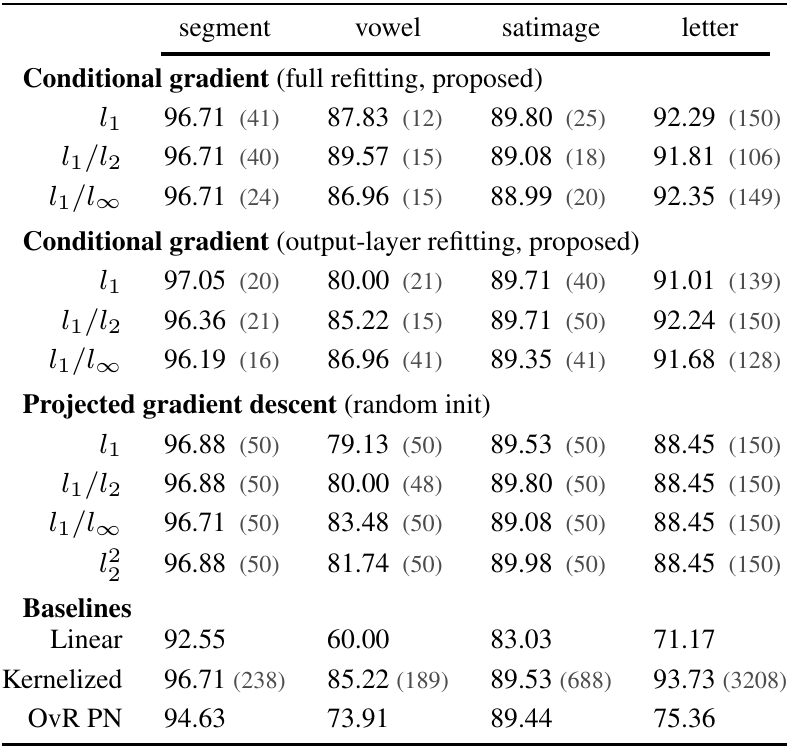}
\label{table:scores}
\end{table}
{\bf Results} are included in Table \ref{table:scores}. From these results, we
can make the following observations and conclusions.  When using output-layer
refitting on vowel and letter (two datasets with more than 10 classes),
group-sparsity inducing penalties lead to better test accuracy. This is to be
expected, since these penalties select basis vectors that are useful across all
classes.  When using full hidden layer and output layer refitting, $l_1$ catches
up with $l_1/l_2$ and $l_1/l_\infty$ on the vowel and letter datasets.
Intuitively, the basis vector selection becomes less important if we make more
effort at every iteration by refitting the basis vectors themselves.  However, on
vowel, $l_1/l_2$ is still substantially better than $l_1$ (89.57 vs.  87.83).

Compared to projected gradient descent with random initialization, our algorithm
(for both output and full refitting) is better on \nicefrac{3}{4} ($l_1$),
\nicefrac{2}{4} ($l_1/l_2$) and \nicefrac{3}{4} ($l_1/l_\infty$) of the
datasets.  In addition, with our algorithm, the best model (chosen against the
validation set) is substantially sparser.  Multi-output PNs substantially
outperform OvR PNs. This is to be expected, since multi-output PNs learn to
share basis vectors across different classes.

\subsection{Recommender system experiments using ordinal regression}

A straightforward way to implement recommender systems consists in training a
single-output model to regress ratings from one-hot encoded user and item
indices \cite{fm}.  
Instead of a single-output  PN or FM, we propose to use ordinal McRank, a
\textbf{reduction from ordinal regression to multi-output binary
classification}, which is known to achieve good nDCG (normalized discounted
cumulative gain) scores \cite{mcrank}.  This reduction involves training a
probabilistic binary classifier for each of the $m$ relevance levels (for
instance, $m=5$ in the MovieLens datasets).  The expected relevance of $\bs{x}$
(e.g.\ the concatenation of the one-hot encoded user and item indices) is then
computed by
\begin{equation}
\small
\hat{y} = \sum_{c=1}^m c ~ p(y=c \mid \bs{x}) = \sum_{c=1}^m c \Big[p(y \le c
    \mid \bs{x}) - p(y \le c-1 \mid \bs{x})\Big],
\end{equation}
where we use the convention $p(y \le 0 \mid \bs{x})=0$. Thus, all we need to do
to use ordinal McRank
is to train a probabilistic binary classifier $p(y \le c \mid \bs{x})$ for all $c
\in [m]$. 

Our key proposal is to use a multi-output model to learn all $m$ classifiers
simultaneously, i.e., in a multi-task fashion. Let
$\bs{x}_i$ be a vector representing a user-item pair with corresponding rating
$y_i$, for $i \in [n]$. We form a $n
\times m$ matrix $\bs{Y}$ such that $y_{i,c} = +1$ if $y_i \le c$ and $-1$
otherwise, and solve
\begin{equation}
\min_{\Omega(\bs{U}) \le \tau} 
\sum_{i=1}^n \sum_{c=1}^m \ell\left(y_{i,c}, 
\sum_{\bs{h} \in \mathcal{H}} \sigma_\text{ANOVA}(\bs{h}, \bs{x}_i)
u_{\bs{h},c} \right),
\end{equation}
where $\ell$ is set to the binary logistic loss, in order to be able to produce
probabilities. After running Algorithm 1 on that objective for $k$ iterations,
we obtain $\bs{H} \in \mathbb{R}^{k \times d}$ and $\bs{V} \in \mathbb{R}^{k
\times m}$. Because $\bs{H}$ is shared across all outputs, the only small
overhead of using the ordinal McRank reduction, compared to a single-output
regression model, therefore comes from learning $\bs{V} \in \mathbb{R}^{k \times
m}$ instead of $\bs{v} \in \mathbb{R}^k$.

\begin{figure}[t]
\centering
\includegraphics[width=0.9\textwidth]{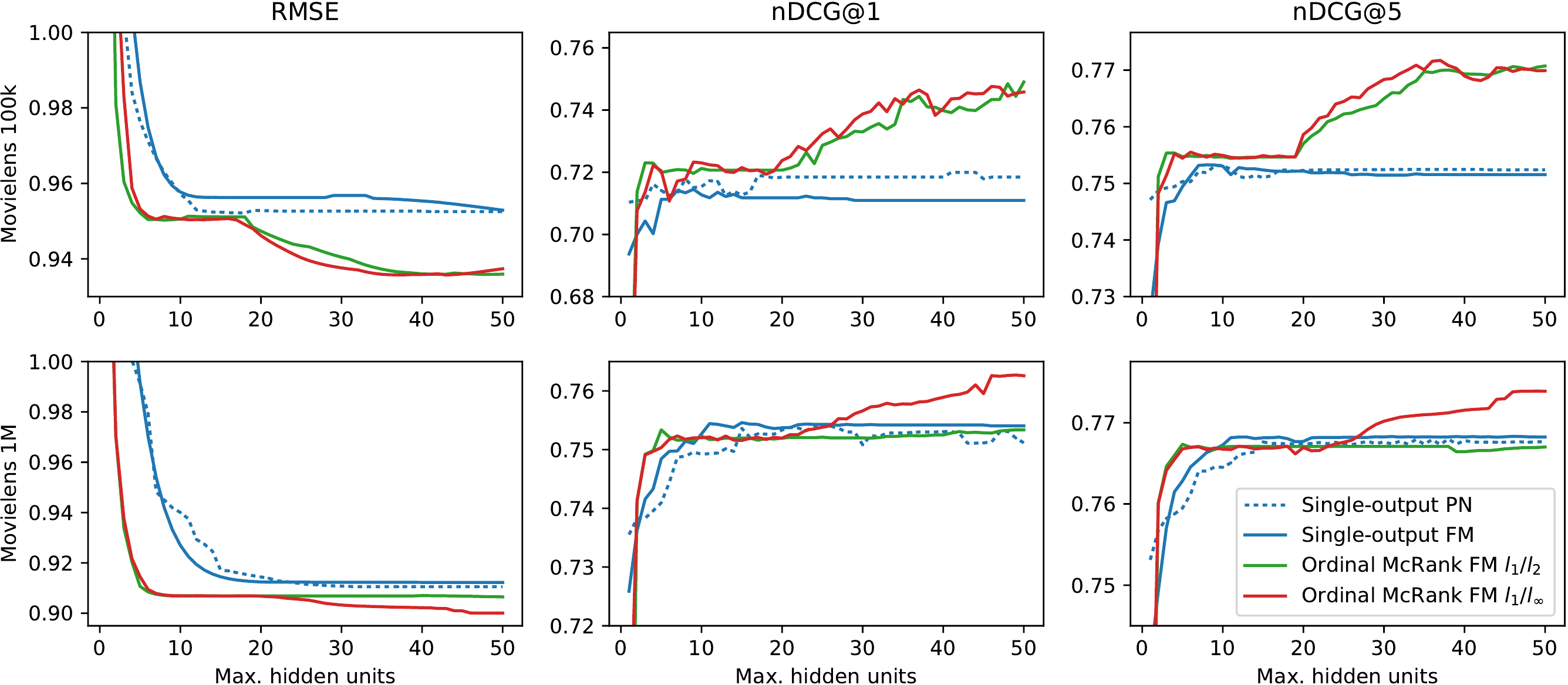}
\caption{Recommender system experiment: RMSE (lower is better) and nDCG
(higher is better).}
\label{fig:recsys_main}
\end{figure}

In this experiment, we focus on multi-output factorization
machines (FMs), since FMs usually work better than PNs for one-hot encoded data
\citep{fm_icml}. We show in Figure \ref{fig:recsys_main} the RMSE and nDCG
(truncated at 1 and 5) achieved when varying
$k$ (the maximum number of basis vectors / hidden units).

{\bf Results.} When combined with the ordinal McRank reduction, we found that
$l_1/l_2$ and $l_1/l_\infty$--constrained multi-output FMs substantially
outperform single-output FMs and PNs on both RMSE and nDCG measures. For
instance, on MovieLens 100k and 1M, $l_1/l_\infty$--constrained multi-output FMs
achieve an nDCG@1 of 0.75 and 0.76, respectively, while single-output FMs only
achieve 0.71 and 0.75. Similar trends are observed with nDCG@5. We believe that
this reduction is more robust to ranking performance measures such as nDCG
thanks to its modelling of the expected relevance.

\section{Conclusion and future directions}

We defined the problem of learning multi-output PNs and FMs as
that of learning a 3-way tensor whose slices share a common basis. To
obtain a convex optimization objective, we reformulated that problem as that of
learning an infinite but row-wise sparse matrix. To learn that matrix, we
developed a conditional gradient algorithm with corrective refitting, and were
able to provide convergence guarantees, despite the non-convexity of the basis
vector (hidden unit) selection step.  

Although not considered in this paper, our algorithm and its analysis can be
modified to make use of stochastic gradients.
An open question remains whether a conditional gradient
algorithm with provable guarantees can be developed for training deep polynomial
networks or factorization machines. Such deep models could potentially
represent high-degree polynomials with few basis vectors.  However, this
would require the introduction of a new functional analysis framework.

\bibliographystyle{abbrv}

\clearpage
\appendix

\begin{center}
    {\Huge \bf Supplementary material}
\end{center}

\section{Convex multi-class loss functions}
\label{appendix:loss_func}

\begin{table}[H]
\caption{ Examples of convex multi-class loss functions $\ell(y, \bs{o}) \in
\mathbb{R}$, where $y \in [m]$ is the correct label and $\bs{o} \in
\mathbb{R}^m$ is a vector of predicted outputs.}
\label{table:loss_func}
\begin{center}
\begin{footnotesize}
\begin{tabular}{l c c}
\toprule
Loss & $\ell(y, \bs{o})$ & $\rho_c(y, \bs{o})$ \\
\midrule
\addlinespace[0.5em]
Multi-class logistic & $\log(1 + \sum_{c \neq y} \exp(o_c -
o_y))$ & 
$\frac{\exp(o_c - o_y)}{\sum_{l=1}^m \exp(o_l - o_y)}$ \\
\addlinespace[0.7em]

Smoothed multi-class hinge  & $\log(1 + \sum_{c \neq y} \exp(1
+ o_c - o_y))$ & 
$\frac{\exp(1[c \neq y] + o_c - o_y)}{\sum_{l=1}^m \exp(1[l \neq y] + o_l -
o_y)}$ \\
\addlinespace[0.7em]

Multi-class squared hinge & $\sum_{c \neq y} \max(1 + o_c -
o_y, 0)^2$ & $2 \max(1 + o_c - o_y, 0)$ \\

\bottomrule
\end{tabular}
\end{footnotesize}
\end{center}
\end{table}

The gradient w.r.t. $\bs{o}$, denoted $\nabla \ell(y, \bs{o}) \in \mathbb{R}^m$,
can be computed by 
\begin{equation}
\nabla \ell(y, \bs{o}) = \sum_{c \neq y} \rho_c(y, \bs{o}) (\bs{e}_c -
\bs{e}_y),
\end{equation}
where $\bs{e}_c \in \mathbb{R}^m$ is a vector whose $c^{\text{th}}$
element is 1 and other elements are 0.  For the smoothed multi-class hinge loss
and the multi-class squared hinge loss, see \cite{share_boost} and
\cite{mblondel_mlj}, respectively.

\section{Proofs}

\subsection{Finite support of an optimal solution (Proposition
\ref{proposition:support})}
\label{appendix:support}

{\bf General case.} We first state a result that holds for arbitrary activation
function $\sigma$ (sigmoid, ReLu, etc...). The main idea is to use the fact that
the penalties considered in Table \ref{table:norms} are atomic
\cite{atomic_norms}. Then, we can equivalently optimize \eqref{eq:convex_obj}
over the convex hull of a set of atoms and invoke Carath\'{e}odory's theorem for
convex hulls.

Let $\phi_{\bs{h}}(\bs{X})$ be an $n$-dimensional vector whose $i^{\text{th}}$
element is $\sigma(\bs{h}^\tr \bs{x}_i)$.  Let us define the sets
\begin{equation}
\mathcal{A} \coloneqq \{\bs{e}_{\bs{h}} \bs{v}^\tr \colon \bs{h} \in
    \mathcal{H}, \bs{v} \in \mathcal{V} \} \subset \RHm
\quad \text{and} \quad
\mathcal{B} \coloneqq \{\phi_{\bs{h}}(\bs{X}) \bs{v}^\tr \colon \bs{h} \in
\mathcal{H}, \bs{v} \in \mathcal{V} \} \subset \mathbb{R}^{n \times m},
\end{equation}
where we define the set $\mathcal{V}$ as follows:
\begin{itemize}
\item $l_1$ case: $\mathcal{V} \coloneqq \{s ~ \bs{e}_c \colon s \in \{-1,1\}, c
    \in [m]\}$
\item $l_1/l_2$ case: $\mathcal{V} \coloneqq \{ \bs{v} \in \mathbb{R}^m
    \colon \|\bs{v}\|_2=1\}$
\item $l_1/l_\infty$ case: $\mathcal{V} \coloneqq \{-1,1\}^m$.
\end{itemize}
Then \eqref{eq:convex_obj} is equivalent to
\begin{equation}
\begin{aligned}
    &\min_{\bs{U} \in \RHm} &&\sum_{i=1}^n \ell\left(y_i,
\sum_{\bs{h} \in \mathcal{H}} \phi_{\bs{h}}(\bs{X})_i ~
\bs{u}_{\bs{h}}\right)
\quad &&\text{s.t.} \quad \Omega(\bs{U}) \le \tau \\
=&\min_{\bs{U} \in \RHm} &&\sum_{i=1}^n \ell\left(y_i,
\sum_{\bs{h} \in \mathcal{H}} \phi_{\bs{h}}(\bs{X})_i ~
\bs{u}_{\bs{h}}\right)
\quad &&\text{s.t.} \quad \bs{U} \in \tau
    \cdot \conv(\mathcal{A}) \\
    =&\min_{\bs{O} \in \mathbb{R}^{n \times m}} &&\sum_{i=1}^n \ell\left(y_i,
        \bs{o}_i \right)
        \quad &&\text{s.t.} \quad \bs{O} \in \tau
    \cdot \conv(\mathcal{B}),
\end{aligned}
\end{equation}
where $\conv(\mathcal{S})$ is the convex hull of the set $\mathcal{S}$.
The matrices $\bs{U}$ and $\bs{O}$ are related to each other by
\begin{equation}
\bs{U} = \sum_{\bs{h} \in \mathcal{H}} \sum_{\bs{v} \in \mathcal{V}}
\theta_{\bs{h},\bs{v}} \bs{e}_{\bs{h}} \bs{v}^\tr
\quad \text{and} \quad
\bs{O} = \sum_{\bs{h} \in \mathcal{H}} \sum_{\bs{v} \in \mathcal{V}}
\theta_{\bs{h},\bs{v}} \phi_{\bs{h}}(\bs{X}) \bs{v}^\tr,
\end{equation}
for some $\bs{\theta} \in \RHm$ such that $\theta_{\bs{h},\bs{v}} \ge 0$
$\forall \bs{h} \in \mathcal{H}, \forall \bs{v} \in \mathcal{V}$ and
$\sum_{\bs{h} \in \mathcal{H}} \sum_{\bs{v} \in \mathcal{V}}
\theta_{\bs{h},\bs{v}} = 1$. By Carath\'{e}odory's theorem for convex hulls,
there exists $\bs{\theta}$ with at most $nm+1$ non-zero elements.
Because elements of $\mathcal{A}$ are matrices with a single non-zero row,
$\bs{U}$ contains at most $nm+1$ non-zero rows (hidden units).

{\bf Case of $l_1$ constraint and squared activation.} When $\sigma(a)=a^2$,
given $\bs{U}$ s.t. $\|\bs{U}\|_1 \le \tau$, the
$c^{\text{th}}$ output can be written as
\begin{equation}
\sum_{\bs{h} \in \mathcal{H}} \sigma(\bs{h}^\tr \bs{x}) u_{\bs{h},c}
= \sum_{\bs{h} \in \mathcal{H}} (\bs{h}^\tr \bs{x})^2 u_{\bs{h},c}
= \bs{x}^\tr \left(\sum_{\bs{h} \in \mathcal{H}} u_{\bs{h},c} \bs{h} \bs{h}^\tr
\right) \bs{x} \eqqcolon \bs{x}^\tr \bs{W}_c \bs{x}.
\end{equation}
Following \cite[Lemma 10]{fm_icml}, the nuclear norm of a symmetric matrix
$\bs{M} \in \mathbb{R}^{d \times d}$ can be defined by
\begin{equation}
    \|\bs{M}\|_* = \underset{\bs{\lambda} \in \mathbb{R}^d, \bs{P} \in
    \mathbb{R}^{d \times d}}{\min}
\sum_{j=1}^d |\lambda_j| ~ \|\bs{p}_j\|^2_2
\quad \text{s.t.} \quad
\bs{M} = \sum_{j=1}^d \lambda_j \bs{p}_j \bs{p}_j^\tr
\label{eq:nuclear_variational_sym}
\end{equation}
and the minimum is attained by the eigendecomposition $\bs{M}=\sum_{j=1}^d
\lambda_j \bs{p}_j \bs{p}_j^\tr$ and $\|\bs{M}\|_* = \|\bs{\lambda}\|_1$. 

Therefore, we can always compute the eigendecomposition of each $\bs{W}_c$ and use
the eigenvectors as hidden units and the eigenvalues as output layer weights.
Moreover, this solution is feasible, since eigenvectors belong to
$\mathcal{H}$ and since the $l_1$ norm of all eigenvalues is minimized.
Since a matrix can have at most $d$ eigenvalues, we can conclude that
$\bs{U}$ has at most $dm$ elements.  Combined with the previous result, $\bs{U}$
has at most $\min(nm+1,dm)$ non-zero rows (hidden units).

For the $l_1/l_2$ and $l_1/l_\infty$ penalties, we cannot make this argument, since
applying the eigendecomposition might increase the penalty value and therefore
make the solution infeasible.

\subsection{Convergence analysis (Theorem \ref{theorem:convergence})}
\label{appendix:convergence_analysis}

In this section, we include a convergence analysis of the
conditional gradient algorithm with multiplicative approximation in the linear
minimization oracle. The proof follows mostly from \cite{revisiting_fw} with a
trick inspired from \cite{convex_nn_bach} to handle multiplicative
approximations.  Finally, we also include a detailed comparison with the
analysis of GECO \cite{geco} and Block-FW \cite{block_fw}.

We focus on constrained optimization problems of the form
\begin{equation}
    \min_{\bs{x} \in \mathcal{D}} f(\bs{x}),
\end{equation}
where $f$ is convex and $\beta$-smooth w.r.t. $\Omega$ and $\mathcal{D}
\coloneqq \{\bs{x}:\Omega(\bs{x}) \le \tau\}$.  

{\bf Curvature and smoothness constants.}
The convergence analysis depends on the following standard curvature constant
\begin{equation}
    \CfD \coloneqq \sup_{\substack{\bs{x}, \bs{s} \in \mathcal{D}\\\gamma \in
        [0,1]\\\bs{y}=\bs{x}+\gamma(\bs{s}-\bs{x})}} \frac{2}{\gamma^2}
        \left(f(\bs{y})-f(\bs{x}) - \langle \bs{y}-\bs{x}, \nabla f(\bs{x})
        \rangle\right).
\end{equation}
Intuitively, this is a measure of non-linearity of $f$: the maximum deviation
between $f$ and its linear approximations over $\mathcal{D}$. The assumption of
bounded $\CfD$ is closely related to a smoothness assumption on $f$.
Following \cite[Lemma 7]{revisiting_fw}, for any choice of norm $\Omega$,
$\CfD$ can be upper-bounded by the smoothness constant $\beta$ as
\begin{equation}
    \CfD \le \text{diam}_\Omega(\mathcal{D})^2 \beta.
\label{eq:curvature_const}
\end{equation}
Using $\mathcal{D}=\{\bs{x}:\Omega(\bs{x}) \le \tau\}$, we obtain
\begin{equation}
    \text{diam}_\Omega(\mathcal{D}) = 
\sup_{\bs{x},\bs{y} \in \mathcal{D}} \Omega(\bs{x}-\bs{y}) \le
\sup_{\bs{x},\bs{y} \in \mathcal{D}} \Omega(\bs{x}) + \Omega(\bs{y}) \le 2 \tau
\end{equation}
and therefore
\begin{equation}
    \CfD \le 4 \tau^2 \beta.
\label{eq:curvature_const_bound}
\end{equation}

{\bf Linear duality gap.}
Following \cite{revisiting_fw}, we define the linear duality gap
\begin{equation}
    g_{\mathcal{D}}(\bs{x}) \coloneqq \max_{\bs{s} \in \mathcal{D}} \langle \bs{x} - \bs{s},
    \nabla f(\bs{x}) \rangle.
\end{equation}

Since $f$ is convex and differentiable, we have that
\begin{equation}
    f(\bs{s}) \ge f(\bs{x}) + \langle \bs{s} - \bs{x}, \nabla f(\bs{x}) \rangle.
\label{eq:convex_inequality}
\end{equation}
Let us define the primal error
\begin{equation}
h_{\mathcal{D}}(\bs{x}) \coloneqq f(\bs{x}) - \min_{\bs{x} \in \mathcal{D}}
f(\bs{x}).
\end{equation}
Minimizing \eqref{eq:convex_inequality} w.r.t. $\bs{s} \in \mathcal{D}$ on both
sides we obtain
\begin{equation}
g_{\mathcal{D}}(\bs{x}) \ge h_{\mathcal{D}}(\bs{x}).
\end{equation}
Hence $g_{\mathcal{D}}(\bs{x})$ can be used as a certificate of optimality about $\bs{x}$.

{\bf Bounding progress.}
Let $\bs{x} \in \mathcal{D}$ be the current iterate and $\bs{y} = \bs{x} + \gamma (\bs{s} -
\bs{x})$ be our update. The definition of $\CfD$ implies 
\begin{equation}
f(\bs{y}) \le f(\bs{x}) + \gamma \langle \bs{s} - \bs{x}, \nabla f(\bs{x}) \rangle +
\frac{\gamma^2}{2} \CfD.
\end{equation}
We now use that $\bs{s}$ is obtained by an exact linear minimization oracle
(LMO)
\begin{equation}
\bs{s} = \argmin_{\bs{s} \in \mathcal{D}} \langle \bs{s}, \nabla f(\bs{x}) \rangle
\end{equation}
and therefore $\langle \bs{s} - \bs{x}, \nabla f(\bs{x}) \rangle =
-g_{\mathcal{D}}(\bs{x})$.
Combined with $g_{\mathcal{D}}(\bs{x}) \ge h_{\mathcal{D}}(\bs{x})$, we obtain
\begin{equation}
f(\bs{y}) \le f(\bs{x}) - \gamma h_{\mathcal{D}}(\bs{x}) + \frac{\gamma^2}{2}
\CfD.
\end{equation}
Subtracting $\min_{\bs{x} \in \mathcal{D}} f(\bs{x})$ on both sides, we finally
get
\begin{equation}
h_{\mathcal{D}}(\bs{y}) \le (1 - \gamma) h_{\mathcal{D}}(\bs{x}) +
\frac{\gamma^2}{2} \CfD.
\end{equation}

{\bf Primal convergence.} Since we use a fully-corrective variant of the
conditional gradient method, our algorithm enjoys a convergence rate at least as
good as the variant with fixed step size.  Following \cite[Theorem
1]{revisiting_fw} and using \eqref{eq:curvature_const_bound}, for every $t \ge
1$, the iterates satisfy
\begin{equation}
f(\iter{x}{t}) - \min_{\bs{x} \in \mathcal{D}} f(\bs{x}) \le \frac{2 \CfD}{t +
2} \le \frac{8 \tau^2 \beta}{t + 2}.
\end{equation}
Thus, we can obtain an $\epsilon$-accurate solution if we run the algorithm for
$t \ge \frac{8 \tau^2 \beta}{\epsilon} - 2$ iterations.

{\bf Linear minimization with multiplicative approximation.}
We now extend the analysis to the case of approximate linear minimization.
Given $\bs{x} \in \mathcal{D}$, we assume that an approximate LMO 
outputs a certain $\bs{s} \in \mathcal{D}$ such that
\begin{equation}
    \langle -\bs{s}, \nabla f(\bs{x}) \rangle \ge
    \nu \max_{\bs{s}' \in \mathcal{D}}
    \langle -\bs{s}', \nabla f(\bs{x}) \rangle,
\end{equation}
for some multiplicative factor $\nu \in (0,1]$ (higher is more accurate).  Since
$\bs{x}$ and $\bs{y} = \bs{x} + \gamma (\bs{s} - \bs{x})$ are in $\mathcal{D}$,
we have like before
\begin{equation}
f(\bs{y}) \le f(\bs{x}) + \gamma \langle \bs{s} - \bs{x}, \nabla f(\bs{x}) \rangle +
\frac{\gamma^2}{2} \CfD.
\end{equation}
Following the same trick as \cite[Appendix B]{convex_nn_bach}, we now absorb
the multiplicative factor $\nu$ in the constraint
\begin{equation}
    \langle -\bs{s}, \nabla f(\bs{x}) \rangle \ge
    \max_{\bs{s}' \in \mathcal{D}'}
    \langle -\bs{s}', \nabla f(\bs{x}) \rangle,
\end{equation}
where we defined $\mathcal{D}' \coloneqq \{\bs{x}:\Omega(\bs{x}) \le \tau \nu\}
= \nu \mathcal{D}$ (i.e., the ball is shrunk by a factor $\nu$). We therefore
obtain $\langle \bs{s} - \bs{x}, \nabla f(\bs{x}) \rangle \le
-g_{\mathcal{D}'}(\bs{x})$.  Similarly as before, this implies that
\begin{equation}
f(\bs{y}) \le f(\bs{x}) - \gamma h_{\mathcal{D}'}(\bs{x}) + \frac{\gamma^2}{2}
\CfD.
\end{equation}
Subtracting $\min_{\bs{x} \in \mathcal{D}'} f(\bs{x})$ on both sides, we get
\begin{equation}
h_{\mathcal{D}'}(\bs{y}) \le (1 - \gamma) h_{\mathcal{D}'}(\bs{x}) +
\frac{\gamma^2}{2} \CfD.
\end{equation}
We thus get that iterate $\iter{x}{t}$ satisfies $\iter{x}{t} \in \mathcal{D}$
and
\begin{equation}
f(\iter{x}{t}) \le \min_{\bs{x} \in \mathcal{D}'} f(\bs{x}) + \frac{8 \tau^2
\beta}{t + 2}.
\end{equation}
We can therefore obtain an $\iter{x}{t} \in \mathcal{D}$ such that
$f(\iter{x}{t}) -
\min_{\bs{x} \in \mathcal{D}'} f(\bs{x}) \le \epsilon$ if we run our algorithm
for $t \ge \frac{8 \tau^2 \beta}{\epsilon} - 2$ iterations with constraint
parameter $\tau$ and multiplicative factor $\nu$.  Put differently, we can
obtain an $\iter{x}{t} \in \frac{1}{\nu}\mathcal{D}$ such that $f(\iter{x}{t}) -
\min_{\bs{x} \in \mathcal{D}} f(\bs{x}) \le \epsilon$ if we run our algorithm
for $t \ge \frac{8 \tau^2 \beta}{\epsilon \nu^2} - 2$ iterations with constraint
parameter $\frac{\tau}{\nu}$ and multiplicative factor $\nu$.

{\bf Comparison with the analysis of GECO.} 
GECO \cite{geco} is a greedy algorithm with fully-corrective
refitting steps for learning a sparse vector from possibly infinitely-many
features, similarly to our algorithm. However, unlike our algorithm, GECO
does not constrain the norm of its iterates (i.e., there is no
parameter $\tau$), which can lead to severe overfitting in practice.
Following \cite[Theorem 1]{geco}, GECO obtains a certain $\iter{x}{t}$
(unbounded) such that
\begin{equation}
    f(\iter{x}{t}) - f(\bs{x}) \le \epsilon \quad 
\forall \bs{x},
\forall t \ge \frac{2 \|\bs{x}\|_1^2 \beta}{\epsilon \nu^2} - 1.
\label{eq:geco_bound}
\end{equation}
In comparison, for the $l_1$-constrained case, our algorithm learns an
$\iter{x}{t}$ such that $\|\iter{x}{t}\|_1 \le \frac{\tau}{\nu}$ and 
\begin{equation}
    f(\iter{x}{t}) - \min_{\|\bs{x}\|_1 \le \tau} f(\bs{x}) \le \epsilon
\quad \forall
t \ge \frac{8 \tau^2 \beta}{\epsilon \nu^2} - 2.
\end{equation}
We see that our algorithm and GECO have similar guarantees, with the difference
that GECO does not constrain its iterates.

GECO was used to learn single-output polynomial networks in \cite{livni}.
Combining \eqref{eq:geco_bound} together with $\|\bs{x}\|_\infty \|\bs{x}\|_0
\ge \|\bs{x}\|_1$, it was shown that GECO can learn the parameters $\iter{x}{t}$
(unbounded) of a single-output polynomial network with $l_\infty$ unit ball
constraint and squared activation such that
\begin{equation}
    f(\iter{x}{t}) - \min_{\|\bs{x}\|_\infty \le 1} f(\bs{x}) \le \epsilon \quad 
\forall \bs{x},
\forall t \ge \frac{2 \|\bs{x}\|_0^2 \beta}{\epsilon \nu^2} - 1.
\end{equation}
However, if we run our algorithm with an $l_1$ constraint, it can learn
an $\iter{x}{t}$ such that $\|\iter{x}{t}\|_1 \le \frac{1}{\nu}$ and
\begin{equation}
    f(\iter{x}{t}) - \min_{\|\bs{x}\|_\infty \le 1} f(\bs{x}) \le
    f(\iter{x}{t}) - \min_{\|\bs{x}\|_1 \le 1} f(\bs{x}) \le \epsilon
\quad \forall
t \ge \frac{8 \beta}{\epsilon \nu^2} - 2.
\end{equation}
Clearly, our algorithm with an $l_1$ constraint uses fewer iterations than GECO
for learning polynomial networks with $l_\infty$ unit ball constraint and more
than $\|\bs{x}\|_0=3$ hidden units.


{\bf Comparison with the analysis of Block-FW.}
\cite{block_fw} analyze a block Frank-Wolfe method with ``multiplicative''
approximations in the linear minimization oracle. However, they require a
different condition, namely:
\begin{align}
&\langle \bs{x} - \bs{s}, \nabla f(\bs{x}) \rangle \ge \kappa \cdot \max_{\bs{s}' \in
\mathcal{D}} \langle \bs{x} - \bs{s}', \nabla f(\bs{x}) \rangle \\
\Leftrightarrow&
\langle -\bs{s}, \nabla f(\bs{x}) \rangle \ge \kappa \cdot \max_{\bs{s}' \in
\mathcal{D}} \langle -\bs{s}', \nabla f(\bs{x}) \rangle + \langle \bs{x}, \nabla
f(\bs{x}) \rangle (\kappa - 1),
\end{align}
for some $\kappa \in (0, 1]$.  Under this condition, they show that the
algorithm converges to an $\epsilon$-approximate solution in
$\mathcal{O}(\frac{1}{\epsilon})$ iterations. A disadvantage of the above
condition is that it contains an additive term that depends on the current
iterate $\bs{x}$ and so it is difficult to give guarantees on $\kappa$ in
general.

\section{Computing an optimal solution of the linearized subproblem
($l_1/l_\infty$ case)}
\label{appendix:l1linf_subproblem}

We describe how to compute an optimal hidden unit $\bs{h}^\star$ in the
$l_1/l_\infty$ case, albeit with exponential complexity in $m$. 
Because of its exponential complexity in $m$
(the number of outputs), clearly, this method should only be used to evaluate
other (polynomial-time) algorithms.

Recall that we want to solve 
\begin{equation}
\max_{\bs{h} \in \mathcal{H}} f_1(\bs{h}) = \sum_{c=1}^m
|\bs{h}^\tr \bs{\Gamma}_c \bs{h}|.
\end{equation}
Now, if we knew the sign $s_c \coloneqq \sign(\bs{h}^{\star \tr} \bs{\Gamma}_c
\bs{h}^\star)$, we could rewrite the problem as
\begin{equation}
\max_{\bs{h} \in \mathcal{H}} f_1(\bs{h}) = \sum_{c=1}^m
s_c \bs{h}^\tr \bs{\Gamma}_c \bs{h} = \bs{h}^\tr \left(\sum_{c=1}^m s_c
\bs{\Gamma}_c \right) \bs{h},
\end{equation}
whose optimal solution is the dominant eigenvector of the symmetric matrix
$\sum_{c=1}^m s_c \bs{\Gamma}_c$. The idea is then simply to find the dominant
eigenvector for all possible $2^m$ sign vectors and choose the eigenvector that
achieves largest objective value. 

\section{Implementation details}
\label{appendix:impl_detail}

In practice, penalized formulations are more convenient to handle than
constrained ones. Here, we discuss why we can safely replace
constrained formulations by penalized formulations in the refitting step.  We
use the output layer refitting objective as an example.
It is well known that there exists $\lambda > 0$ such that this objective is
equivalent to
\begin{equation}
\min_{\bs{V} \in \mathbb{R}^{t \times m}}
F(\bs{V}, \iter{H}{t}) + \lambda \Omega(\bs{V}). 
\end{equation}
Unfortunately, the relation between $\tau$ and $\lambda$ is a priori unknown.
However, it is easy to see that the constant factor $\tau$ in
\eqref{eq:linearized_pb} is absorbed by the output layer in our refitting
step. This means that we need to know the actual value of $\tau$ for the
refitting step but not for the hidden unit selection step. As long as we
compute a full regularization path, we may therefore use a penalized formulation
in a practical implementation. We do so for both refitting objectives we
discussed.

For both refitting objectives,
we use FISTA, an accelerated projected gradient method with
$\mathcal{O}(1/t^2)$ convergence rate, where $t$ is the iteration number.
We set the maximum number of iterations to $1000$ and the stopping criterion's
tolerance to $10^{-3}$. 

\section{Datasets}
\label{appendix:datasets}

For our multi-class experiments, we used the following four publicly available
datasets \cite{datasets}.
\begin{table}[H]
\label{table:datasets}
\begin{center}
\begin{tabular}{l c c c}
\toprule
Name & $n$ & $d$ & $m$ \\
\midrule
\addlinespace[0.5em]
segment & 2,310 & 19 & 7 \\
\addlinespace[0.7em]
vowel & 528 & 10 & 11 \\
\addlinespace[0.7em]
satimage & 4,435 & 36 & 6 \\
\addlinespace[0.7em]
letter & 15,000 & 16 & 26 \\

\bottomrule
\end{tabular}
\end{center}
\end{table}

For recommender system experiments, we used the following two publicly available
datasets \cite{movielens}.
\begin{table}[H]
\label{table:datasets_recsys}
\begin{center}
\begin{tabular}{l c c c}
\toprule
Name & $n$ & $d$ & $m$\\
\midrule
\addlinespace[0.5em]
Movielens 100k & 100,000 (ratings) & 2,625 = 943 (users) + 1,682 (movies) & 5\\
\addlinespace[0.5em]
Movielens 1M & 1,000,209 (ratings) & 9,940 = 6,040 (users) + 3,900 (movies) & 5\\

\bottomrule
\end{tabular}
\end{center}
\end{table}

The task is to predict ratings between 1 and 5 given by users to movies, i.e.,
$y \in \{1, \dots, 5\}$.  The design matrix $\bs{X}$ was constructed following
\cite{fm}. Namely, for each rating $y_i$, the corresponding $\bs{x}_i$ is set
to the concatenation of the one-hot encodings of the user and item indices.
Hence the number of samples $n$ is the number of ratings and the number of
features is equal to the sum of the number of users and items. Each sample
contains exactly two non-zero features. It is known that factorization machines
are equivalent to matrix factorization when using this representation
\cite{fm}.

\section{Additional experimental results}

\subsection{Multi-class squared hinge loss results}

We also compared the multi-class logistic (ML) loss to the multi-class squared
hinge (MSH) loss. The MSH loss achieves comparable test accuracy to the ML loss.
However, it can often be much faster to train, since it does not require
expensive exponential and logarithm calculations.

\begin{table}[H]
    \caption{Comparison betwen multi-class squared hinge (MSH) and
    logistic (ML) losses.}
\vspace{0.3cm}
\fontsize{7}{8}\selectfont
\centering
\begin{tabular}{r C{1cm} C{1cm} C{1cm} C{1cm} C{1cm} C{1cm}
C{1cm} C{1cm} C{1cm}}
\toprule
\multirow{2}{*}[-2pt]{Constraint} & \multirow{2}{*}[-2pt]{Loss} &
\multicolumn{4}{c}{{\bfseries Conditional gradient} (full refitting)} &
\multicolumn{4}{c}{{\bfseries Conditional gradient} (output-layer refitting)} \\
\cmidrule[0.5pt](rl){3-6}
\cmidrule[0.5pt](rl){7-10}
 & & segment & vowel & satimage & letter & segment & vowel & satimage & letter\\
\cmidrule[0.8pt](rl){1-2}
\cmidrule[0.8pt](rl){3-6}
\cmidrule[0.8pt](rl){7-10}
& \multirow{2}{*}{MSH} & 96.01 & 87.83 & 89.98 & 92.03 & 95.67 & 79.13 & 88.99 &
91.25 \\
$l_1$ &  & {\color{supgray}({21})} & {\color{supgray}({8})} &
{\color{supgray}({22})} & {\color{supgray}({130})} & {\color{supgray}({21})} &
{\color{supgray}({25})}
& {\color{supgray}({21})} & {\color{supgray}({149})} \\
\addlinespace[0.3em] {\color{supgray} (\#units)}& \multirow{2}{*}{ML}&  96.71 &
87.83 & 89.80 & 92.29 & 97.05 & 80.00 & 89.71 & 91.01 \\
& & {\color{supgray}({41})} & {\color{supgray}({12})}
                            & {\color{supgray}({25})} & {\color{supgray}({150})}
                            & {\color{supgray}({20})} & {\color{supgray}({21})}
                            & {\color{supgray}({40})} & {\color{supgray}({139})}\\
\cmidrule[0.5pt](rl){3-6}
\cmidrule[0.5pt](rl){7-10}
& \multirow{2}{*}{MSH} & 96.01 & 86.96 & 90.25 & 91.57 & 95.67 & 85.22 & 89.98 &
92.03 \\
$l_1/l_2$& & {\color{supgray}({15})} & {\color{supgray}({8})} &
{\color{supgray}({12})} & {\color{supgray}({94})} & {\color{supgray}({25})} &
{\color{supgray}({19})} &
{\color{supgray}({50})} & {\color{supgray}({149})} \\
\addlinespace[0.3em]
{\color{supgray} (\#units)} & \multirow{2}{*}{ML} & 96.71 & 89.57 & 89.08 &
91.81  & 96.36 & 85.22 & 89.71 & 92.24 \\
& & {\color{supgray}({40})} & {\color{supgray}({15})}
                            & {\color{supgray}({18})} & {\color{supgray}({106})}
                            & {\color{supgray}({21})} & {\color{supgray}({15})}
                            & {\color{supgray}({50})} & {\color{supgray}({150})} \\
\cmidrule[0.5pt](rl){3-6}
\cmidrule[0.5pt](rl){7-10}
& \multirow{2}{*}{MSH} & 95.84 & 85.22 & 89.80 & 92.27  & 97.05 & 86.09 & 88.90
& 91.20 \\
$l_1/l_\infty$ & &  {\color{supgray}(16)} & {\color{supgray}(18)} &
{\color{supgray}(29)} & {\color{supgray}(149)}  & {\color{supgray}(28)} &
{\color{supgray}(33)} & {\color{supgray}(24)} & {\color{supgray}(119)} \\
\addlinespace[0.3em]
{\color{supgray}(\#units)} & \multirow{2}{*}{ML} & 96.71 & 86.96 & 88.99 & 92.35
                           & 96.19 & 86.96 & 89.35 & 91.68 \\
 & & {\color{supgray}(24)} & {\color{supgray}(15)} &
{\color{supgray}(20)} & {\color{supgray}(149)} & {\color{supgray}(16)} &
{\color{supgray}(41)} &
{\color{supgray}(41)} & {\color{supgray}(128)} \\
\bottomrule
\end{tabular}
\label{table:sqhingescores}
\end{table}

\subsection{Full vs. output layer refitting comparison}

In this experiment, we compare output layer refitting with full refitting of
both the hidden and output layers.  Empirically, we observe that full refitting
does not always outperform output layer refitting in terms of objective value
but it does so in terms of test accuracy.

\begin{figure}[H]
\makebox[\linewidth]{
        \includegraphics[width=0.9\linewidth]{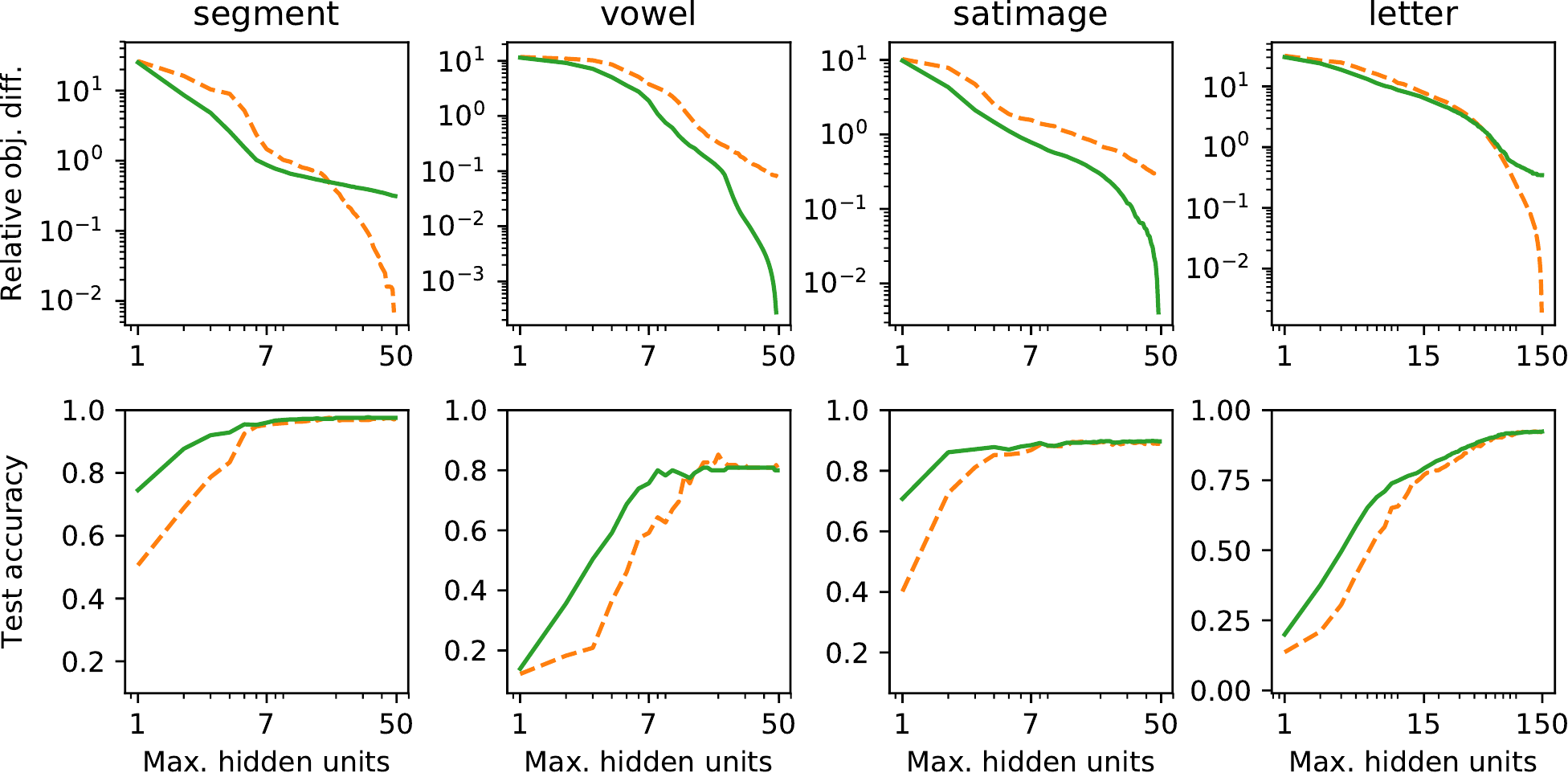}
    }
\caption{Relative objective difference from best (top) and multi-class test set
accuracy values (bottom) when performing output layer refitting (dashed) and
full, non-convex refitting (solid), optimizing a penalized $l_1/l_2$ objective
with $\lambda=0.1$.}
\label{fig:refit}
\end{figure}

\end{document}